\documentclass[letterpaper,10pt,conference]{ieeeconf}
\IEEEoverridecommandlockouts

\usepackage[T1]{fontenc}
\usepackage{graphicx}
\usepackage{tikz}
\usetikzlibrary{arrows.meta,positioning}
\usepackage{amsmath,amssymb}
\usepackage{array}
\usepackage{cite}
\usepackage{balance}
\usepackage{afterpage}
\usepackage[hidelinks]{hyperref}

\hypersetup{
  pdftitle={CoCarry: Contact–Support-Structured Retargeting for Dual-Humanoid Cooperative Carrying},
  pdfauthor={Ziyu Lin, Shihang Du, Tianhang Wang, Xiaodong Liu, Runze Zhang, Zongyou Yu, Haodong Zhang, Alois Knoll, Guang Chen},
  pdfsubject={},
  pdfkeywords={}
}

\title{\LARGE \bf
CoCarry: Contact–Support-Structured Retargeting for Dual-Humanoid Cooperative Carrying
}

\author{%
  \authorblockN{%
    \textbf{Ziyu Lin}\textsuperscript{1,2,3,4,\textdagger},
    \textbf{Shihang Du}\textsuperscript{1,\textdagger},
    \textbf{Tianhang Wang}\textsuperscript{1},
    \textbf{Xiaodong Liu}\textsuperscript{1},\\
    \textbf{Runze Zhang}\textsuperscript{1},
    \textbf{Zongyou Yu}\textsuperscript{1},
    \textbf{Haodong Zhang}\textsuperscript{1},
    \textbf{Alois Knoll}\textsuperscript{5},
    \textbf{Guang Chen}\textsuperscript{1,2,3,\textdaggerdbl}%
  }%
  \authorblockA{\small
    \textsuperscript{1}Physical Intelligence Lab, Tongji University, Shanghai, China\quad
    \textsuperscript{2}Shanghai Innovation Institute, Shanghai, China\\
    \textsuperscript{3}RobotScale.AI\quad
    \textsuperscript{4}AGIBOT, Shanghai, China\quad
    \textsuperscript{5}Technical University of Munich, Germany\\
    \textsuperscript{\textdagger}Equal contribution.\quad
    \textsuperscript{\textdaggerdbl}Corresponding author.%
  }%
}

\IEEEaftertitletext{%
  \vspace{-0.7em}%
  \centering
  \includegraphics[width=0.93\textwidth]{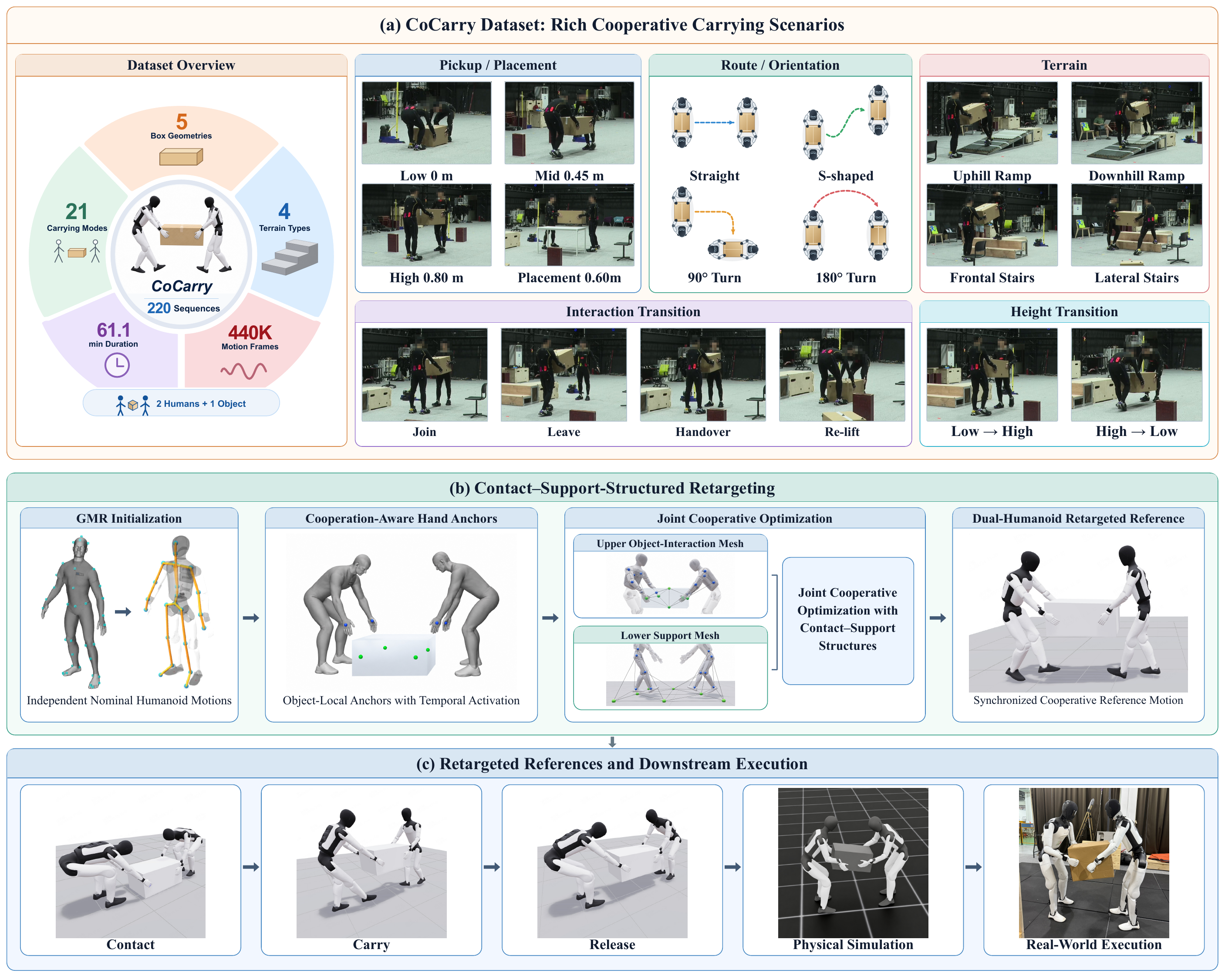}%
  \\[-0.15em]%
  \refstepcounter{figure}%
  \label{fig:cocarry-overview}%
  {\footnotesize
  \textbf{Fig.~\thefigure.}
  Overview of CoCarry and the proposed Contact--Support-Structured Retargeting framework. (a) CoCarry contains synchronized two-person carrying sequences covering diverse task schemes, carrying heights, routes, terrains, and object geometries. (b) Starting from nominal individual retargeting, cooperation-aware hand anchors and complementary upper object-interaction and lower support structures jointly refine both humanoids around the shared object. (c) Representative keyframes illustrate downstream execution in physics simulation and on physical A3 humanoids.
  }%
  \\[0.7em]%
}

\begin{document}

\maketitle
\thispagestyle{empty}
\pagestyle{empty}

\begin{abstract}
Recent advances in large-scale human motion data, motion retargeting, and
motion imitation have enabled increasingly capable whole-body control for
individual humanoids. Cooperative carrying is more challenging because
successful coordination requires not only plausible individual motions, but
also consistent cross-agent and agent--object relations, including shared
grasp, relative configuration, and whole-body support. We introduce
\textbf{CoCarry}, a cooperative-carrying dataset containing 220 synchronized
two-person sequences across 21 task schemes, five shared-object geometries,
four terrain types, and approximately 61 minutes of motion. We further propose
\textbf{Contact--Support-Structured Retargeting}, which starts from nominal
individual motions and jointly refines both humanoids around the same
prescribed object trajectory. Cooperation-aware object-local hand anchors
preserve compatible four-hand grasp relations, while complementary upper
object-interaction and lower support structures retain object-relative
geometry and coordinated support. Across all 220 CoCarry sequences retargeted
to two A3 humanoids, our method achieves 0.9995 contact preservation, reduces
penetration duration by 67.0\% and 71.0\% relative to GMR and OmniRetarget,
respectively, and maintains a 100\% retargeting success rate. On CORE4D with
Unitree G1 humanoids, it achieves 0.9906 contact preservation with zero
measured sliding. We further validate the generated references through
physics simulation and multiple real-world cooperative carrying experiments
on two physical A3 humanoids, demonstrating their usability for downstream
dual-humanoid control.
\end{abstract}\textbf{}

\section{Introduction}
\label{sec:introduction}

Recent advances in large-scale human motion data, motion retargeting, and
motion imitation have substantially improved whole-body control for individual
humanoids \cite{deepmimic,amp,phc,pulse,sonic,intermimic}. Extending these
capabilities to cooperative tasks is more challenging because two participants
must preserve not only their individual motions, but also cross-agent and
agent--object relations. In cooperative carrying, shared grasp, relative
configuration, and whole-body support must remain mutually consistent
throughout the task.

Existing human--human and human--object--human datasets provide important
resources for multi-person interaction and collaborative manipulation
\cite{core4d,hoim3,interx,interxplus,grab,omomo,hoh}. Our work is
complementary to these efforts and focuses specifically on
\textbf{dual-humanoid cooperative carrying}. We introduce \textbf{CoCarry},
a dataset of 220 synchronized two-person carrying sequences with shared-object
trajectories, covering diverse task schemes, object geometries, and terrain
conditions. Each sequence preserves both participants and the shared object on
a common timeline together with time-varying hand--object interaction priors.
Figure~\ref{fig:cocarry-overview} summarizes the dataset, proposed retargeting
framework, and downstream execution.

A key challenge is converting these synchronized human motions into
dual-humanoid references. Existing methods provide strong foundations for
general and interaction-aware retargeting
\cite{gmr,interactionmesh,omniretarget}, but independently retargeting the two
participants does not explicitly enforce relations required by the shared
task. Two individually plausible motions may still produce incompatible hand
placements, distorted relative configurations, penetration, or inconsistent
support. Thus, \textbf{good individual retargeting does not necessarily imply
good cooperative retargeting}.

We therefore propose \textbf{Contact--Support-Structured Retargeting}, which
starts from nominal individual motions and jointly refines both humanoids under
cooperative constraints. Cooperation-aware object-local hand anchors preserve
shared grasp relations, while complementary upper object-interaction and lower
support structures preserve object-relative interaction and ground-referenced
support. Both humanoids are optimized jointly around the same prescribed
object trajectory.

We evaluate the method on all 220 CoCarry sequences with two A3 humanoids,
on a matched CORE4D subset with two Unitree G1 humanoids, and through physics
simulation and real-world cooperative carrying experiments on two physical
A3 humanoids. These evaluations assess retargeting quality, cross-dataset and
cross-embodiment generalization, and downstream physical usability.

Our contributions are:
\begin{itemize}
    \item \textbf{CoCarry}, a cooperative-carrying dataset with synchronized
    two-person motion, shared-object trajectories, and time-varying
    hand--object interaction priors.

    \item \textbf{Contact--Support-Structured Retargeting}, a joint
    dual-humanoid retargeting method for preserving cooperative grasp and
    whole-body support.

    \item \textbf{Comprehensive validation} across A3 and G1 humanoids,
    multiple datasets, simulation, and real-world cooperative execution.
\end{itemize}

\section{Related Work}
\label{sec:related-work}

\subsection{Human Interaction and Cooperative Motion Datasets}

Existing datasets increasingly cover human--human and human--object
interaction. Inter-X and Inter-X++ capture diverse two-person interactions,
GRAB and OMOMO focus on whole-body human--object motion, while HOH, HOI-M3,
and CORE4D further consider handover, multi-human--object interaction, and
collaborative object manipulation
\cite{interx,interxplus,grab,omomo,hoh,hoim3,core4d}.
CoCarry complements these resources by focusing specifically on cooperative
carrying, where two participants continuously coordinate around the same
moving object. It preserves synchronized two-person motion, shared-object
trajectories, diverse carrying conditions, and time-varying hand--object
interaction priors for downstream humanoid retargeting.

\subsection{Humanoid Motion Retargeting}

Human-to-humanoid retargeting converts human motion into robot reference
trajectories. GMR provides a general retargeting framework, while spatial
relationship-preserving adaptation and OmniRetarget further incorporate
interaction, object, and environment relations
\cite{gmr,interactionmesh,omniretarget}. Multi-character methods also study
relation-aware motion transfer and refinement
\cite{multicharacter,synagent,mochi}.

For cooperative carrying, however, two individually plausible motions must
also remain mutually compatible around the same shared object. Even if each
participant is retargeted to a plausible humanoid motion, the resulting pair
may still exhibit mismatched hand placement, distorted relative
configuration, object penetration, or inconsistent support. Preserving two
individual motions therefore does not by itself guarantee preservation of the
cooperative relation between them. Our method uses individual retargeting as
initialization and explicitly preserves shared grasp, relative configuration,
and support relations through joint dual-humanoid refinement.

\subsection{Humanoid Imitation and Cooperative Control}

DeepMimic, AMP, PHC, PULSE, SONIC, and InterMimic have advanced physics-based
humanoid motion imitation and whole-body control
\cite{deepmimic,amp,phc,pulse,sonic,intermimic}. Cooperative object
transportation, human--humanoid carrying, and multi-agent interaction have
also been studied \cite{hawley,cola,coohoi,teamhoi}.

Our focus is complementary and lies upstream of control: we construct
dual-humanoid cooperative references from synchronized human carrying motion,
while motion imitation is used to validate whether these references can
support downstream physical execution.

\section{CoCarry Dataset}
\label{sec:dataset}

\subsection{Dataset Scope and Cooperative Task Coverage}

\textbf{CoCarry} contains \textbf{220 synchronized two-person carrying
sequences}, with approximately \textbf{61.1\,min} of motion and
\textbf{440K raw MoCap frames} recorded at 120\,Hz. Each sequence contains
two human motion streams and one shared-object trajectory on a common
timeline, together with object, scene, and task annotations.

As shown in Fig.~\ref{fig:cocarry-overview}(a), the dataset contains
\textbf{21 carrying schemes} grouped into five task families:
\emph{pickup/placement}, \emph{route/orientation}, \emph{terrain},
\emph{interaction transition}, and \emph{height transition}.

The \emph{pickup/placement} family covers different pickup and carrying
heights together with placement onto a support surface. The
\emph{route/orientation} family includes straight and S-shaped navigation as
well as coordinated $90^{\circ}$ and $180^{\circ}$ turns. The
\emph{terrain} family contains uphill and downhill ramp carrying and frontal
and lateral stair traversal. The \emph{interaction-transition} family changes
how the two participants engage with the shared task, including joining,
leaving, handover, and temporary placement. Finally, the
\emph{height-transition} family captures low-to-high and high-to-low object
motion during carrying.

CoCarry further contains five shared-object geometries with different scales
and aspect ratios. These variations affect the feasible separation between
the two carriers, arm reach, four-hand grasp configuration, and whole-body
support, making the same cooperative task nontrivial across different
geometric conditions.

\begin{figure}[t]
    \centering
    \includegraphics[width=\columnwidth]
    {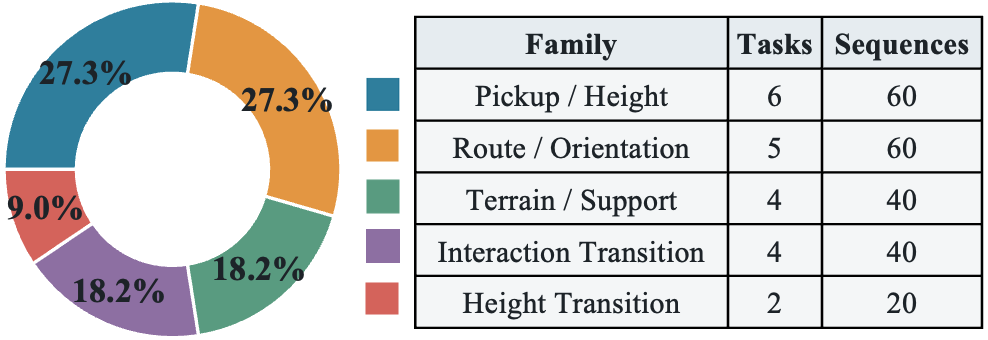}
    \caption{
    Distribution of the 220 CoCarry sequences across five
    cooperative-carrying task families. The table reports the number of task
    schemes and sequences in each family.
    }
    \label{fig:cocarry-distribution}
    \vspace{-1ex}
\end{figure}

Figure~\ref{fig:cocarry-distribution} shows the quantitative distribution of
the dataset. The 21 schemes contain 6 pickup/placement, 5 route/orientation,
4 terrain, 4 interaction-transition, and 2 height-transition tasks,
corresponding to 60, 60, 40, 40, and 20 sequences, respectively.

\subsection{Synchronized Human--Object Representation}

CoCarry represents the two participants and the shared object as one
synchronized interaction sequence rather than independent motion clips. Both
human streams retain their identities and share the same object trajectory,
scene geometry, and task timeline. This preserves not only individual motion
but also the relative placement, shared grasp, and participation changes that
define the cooperative task.

The raw motion is captured at 120\,Hz. For downstream humanoid retargeting,
the two human streams and the shared-object trajectory are resampled to a
common 50\,Hz timeline while maintaining framewise correspondence. This
provides a consistent temporal basis for jointly reasoning about both
participants and the same moving object.

\subsection{Cooperative Interaction Priors}

Beyond synchronized trajectories, CoCarry provides interaction priors that
describe the hand--object relations needed by cooperative retargeting. For
the four hands of the two participants, we derive four object-local anchors.
Each anchor $a_h$ is associated with a surface normal $\nu_h$ and a
time-varying activation variable
\[
    \alpha_{ht}\in[0,1],
\]
which encode the contact location, palm orientation, and active time of the
corresponding hand--object relation.

The temporal activation is important for task schemes in which participation
changes over time. It allows the same representation to describe grasp
establishment, sustained carrying, joining or leaving, handover, temporary
placement, and release without assuming permanent four-hand contact. These
quantities are kinematic priors derived from synchronized human--object motion
rather than force measurements.

Together, the synchronized trajectories, task annotations, and interaction
priors provide the structured inputs used by the
Contact--Support-Structured Retargeting method in
Sec.~\ref{sec:retargeting}. The motion data, object geometry, annotations,
interaction priors, processing tools, and available dual-humanoid references
are planned for release.

\section{Contact--Support-Structured Retargeting}
\label{sec:retargeting}

\begin{figure*}[t]
    \centering
    \includegraphics[width=0.98\textwidth]
    {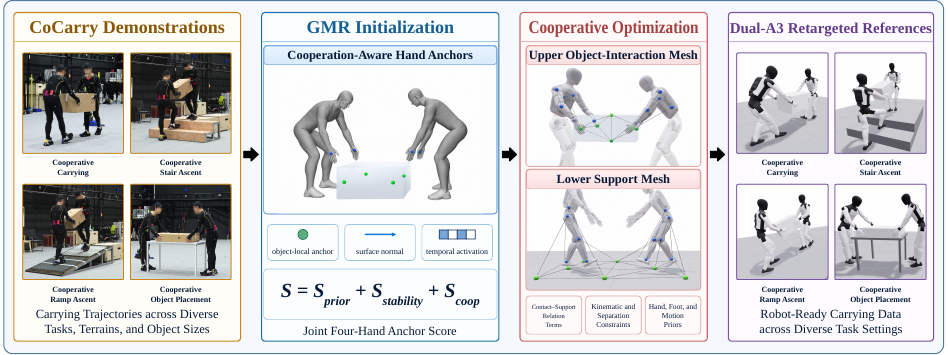}
    \caption{
    Detailed pipeline of Contact--Support-Structured Retargeting.
    Starting from synchronized human motions from CoCarry, GMR provides
    nominal motions for both humanoids. Cooperation-aware hand anchors encode
    shared grasp relations, while complementary upper object-interaction and
    lower support structures guide joint refinement around the prescribed
    object trajectory.
    }
    \label{fig:retargeting-pipeline}
\end{figure*}

\subsection{Problem Definition and GMR Initialization}

At time $t$, CoCarry provides synchronized human motions $H_{1t}$ and $H_{2t}$,
a shared box mesh $B$, and a prescribed object pose $T_t=(R_t,o_t)$. We seek the
stacked humanoid configuration $q_t=(q_{1t},q_{2t})$ that preserves the
cooperative relations around the moving object. The fixed object trajectory
serves as the common reference for object-related interaction terms.

We first apply GMR~\cite{gmr} independently to obtain nominal humanoid motions:
\begin{equation}
  \bar q_{it}=\operatorname{GMR}(H_{it}),
  \qquad i\in\{1,2\}.
  \label{eq:rt-gmr}
\end{equation}
The two nominal motions are synchronized with the object trajectory before
joint refinement. The resulting $\bar q_t=(\bar q_{1t},\bar q_{2t})$ provides
the initialization and nominal-motion prior. Although individually plausible,
independent retargeting does not preserve the shared grasp, relative
configuration, and coordinated support required for cooperative carrying.

For a world-frame point $x_t^W$, object-local coordinates are
\begin{equation}
  x_t^O=R_t^\top\left(x_t^W-o_t\right).
  \label{eq:rt-object-frame}
\end{equation}
Upper-body interaction quantities are evaluated in this object frame while
optimizing the two humanoid configurations with the prescribed object
trajectory fixed.

\subsection{Cooperation-Aware Hand Anchors}

Independent hand tracking cannot determine whether the four hands form a
mutually compatible grasp. We therefore extract four cooperation-aware anchors
from synchronized human--object motion, one for each hand. The anchors are
represented in the object frame so that their relative grasp geometry is
preserved as the object moves.

Let $h^O_{t,h}$ denote the demonstrated position of hand $h$ in the object
frame. For each candidate carrying window $\mathcal{W}$ and feasible
hand--surface configuration, the anchor is obtained from the temporal median
of the hand trajectory and projected onto the assigned finite object surface:
\begin{equation*}
  a_h=
  \Pi_{B,h}
  \left(
  \operatorname*{med}_{t\in\mathcal{W}}
  h^O_{t,h}
  \right),
\end{equation*}
where $\Pi_{B,h}$ projects onto the corresponding box surface. This preserves
the demonstrated longitudinal and vertical contact location within the valid
surface region and associates the anchor with the outward surface normal
$\nu_h$.

The four anchors are evaluated jointly using
\begin{equation*}
  S
  =
  S_{\mathrm{prior}}
  +
  S_{\mathrm{stability}}
  +
  S_{\mathrm{coop}},
\end{equation*}
where
\begin{equation*}
  S_{\mathrm{prior}}
  =
  S_{\mathrm{surf}}
  +
  0.1S_{\mathrm{prior-pos}},
\end{equation*}
\begin{equation*}
  S_{\mathrm{stability}}
  =
  2S_{\mathrm{static}}
  +
  0.05S_{\mathrm{prior-norm}},
\end{equation*}
and
\begin{equation*}
  S_{\mathrm{coop}}
  =
  0.5S_{\mathrm{pair}}
  +
  0.25S_{\mathrm{balance}}.
\end{equation*}

The prior term combines demonstrated hand--surface proximity
$S_{\mathrm{surf}}$ with nominal GMR palm--anchor distance
$S_{\mathrm{prior-pos}}$. The human motion provides the main grasp-location
cue, while the nominal robot motion acts as a weaker feasibility prior. The
stability term favors locally consistent contact: $S_{\mathrm{static}}$
penalizes tangential hand drift, while $S_{\mathrm{prior-norm}}$ measures palm
orientation relative to the contact surface. The cooperation term jointly evaluates the four anchors: $S_{\mathrm{pair}}$ describes each carrier's two-hand layout, while $S_{\mathrm{balance}}$ favors anchor configurations that form a compatible spatial organization between the two carriers. Together, these terms capture demonstrated
hand placement, temporal contact stability, and four-hand compatibility. Joint evaluation discourages
individually plausible anchors that would create an inconsistent cooperative
grasp. The lowest-scoring valid candidate provides the final anchors.

\emph{Anchor Activation.}
Fixed anchors define desired contact regions but should not constrain the hands
throughout the entire sequence. At frame $t$, the world-frame anchor position
is
\begin{equation*}
  a^W_{t,h}=o_t+R_ta_h.
\end{equation*}
We compare it with both the demonstrated hand and the nominal GMR palm:
\begin{equation*}
\begin{aligned}
  d^H_{t,h}
  &=
  \left\|
  h^W_{t,h}-a^W_{t,h}
  \right\|_2,
  \qquad
  d^G_{t,h}
  =
  \left\|
  g^W_{t,h}-a^W_{t,h}
  \right\|_2,
\end{aligned}
\end{equation*}
where $h^W_{t,h}$ and $g^W_{t,h}$ denote the demonstrated hand and nominal GMR
palm positions. The two distance cues jointly determine a framewise activation
mask. During sustained off-ground carrying, each hand is activated according
to its own hand--anchor relation, allowing different grasp participation
states.

Each active interval is converted into a five-frame smoothstep weight
$\alpha_{ht}\in[0,1]$ to avoid abrupt contact establishment or release. The
resulting hand objective is
\begin{equation}
\begin{aligned}
  E_{\mathrm{hand}}(q_t)
  =
  \sum_{h=1}^{4}
  \alpha_{ht}
  \biggl[
  &\lambda_p
  \left\|
  p_h^O(q_t)-a_h
  \right\|_2^2 \\
  {}+{}&
  \lambda_n
  \left\|
  n_h^O(q_t)+\nu_h
  \right\|_2^2
  \biggr].
\end{aligned}
\label{eq:rt-hand}
\end{equation}
Here, $p_h^O(q_t)$ and $n_h^O(q_t)$ denote the palm position and outward palm
normal in the object frame, respectively. Because $\nu_h$ is the outward
object-surface normal,
\begin{equation*}
  n_h^O(q_t)\approx-\nu_h
\end{equation*}
means that the palm faces the contacted surface. The time-varying weights
represent grasp establishment, sustained carrying, transitions, and release
without forcing permanent four-hand contact.

\subsection{Contact--Support Structured Representation}

Hand anchors describe task-critical palm--object relations but not the
whole-body geometry of cooperative carrying. We therefore construct two
complementary relational meshes: an upper object-interaction mesh for
shared-object manipulation and a lower support mesh for coordinated support
relative to the ground.

The upper mesh contains the shoulder, elbow, and hand landmarks of both
carriers together with shared object-surface samples. Human and robot meshes
use matched landmarks and the prescribed object geometry in the object frame.
Shared object nodes preserve arm configurations and coordinated carrier
arrangement while remaining invariant to global object motion, complementing
local palm constraints.

The lower mesh uses both carriers' pelvis, leg, and foot landmarks with ground
reference nodes in the world frame. It preserves coordinated lower-body
configuration and stance during carrying, turning, ramp walking, and stair
motion. Object-relative upper-body and ground-referenced lower-body meshes
therefore capture the distinct frames required by manipulation and locomotion,
while remaining coupled in the same joint dual-humanoid optimization.

For each structure $g\in\{U,L\}$, let $V_H^g$ and $V_R^g(q_t)$ denote the
human and robot landmark sets with matched semantic correspondences and
relational topology, and let $L_g$ denote the graph Laplacian. The
relation-preservation error is
\begin{equation}
  E_g(q_t)
  =
  \left\|
  W_g^{1/2}
  \left(
  L_gV_R^g(q_t)-L_gV_H^g
  \right)
  \right\|_2^2,
  \qquad
  g\in\{U,L\},
  \label{eq:rt-structure}
\end{equation}
where $W_g$ weights selected relations. The upper-body term $E_U$ preserves
cooperative geometry between both carriers and the shared object, whereas the
lower-body term $E_L$ preserves coordinated support relative to the ground.
Active hand anchors further constrain local palm--object contacts. Together,
these terms preserve shared grasp, cross-robot spatial relations, and
lower-body support without explicitly modeling forces, friction, or contact
dynamics.

\subsection{Joint Dual-Humanoid Cooperative Optimization}

At each frame, both humanoids are jointly optimized around the fixed object
pose. Let
\begin{equation*}
  \Delta q=(\Delta q_1,\Delta q_2)
\end{equation*}
denote their stacked local update. The geometric objectives and distance
constraints are locally linearized and solved sequentially:
\begin{equation}
\begin{aligned}
  \min_{\Delta q,\,s\geq 0}\quad
  &
  \lambda_UE_U
  +
  \lambda_LE_L
  +
  E_{\mathrm{hand}}
  +
  E_{\mathrm{foot}}
  \\
  &
  +
  \lambda_{\mathrm{nom}}
  \left\|
  q_t+\Delta q-\bar q_t
  \right\|_2^2
  \\
  &
  +
  \lambda_{\mathrm{tmp}}
  \left\|
  q_t+\Delta q-q_{t-1}
  \right\|_2^2
  +
  \lambda_s\|s\|_2^2
  \\[2pt]
  \text{s.t.}\quad
  &
  q_{\min}
  \leq
  q_t+\Delta q
  \leq
  q_{\max},
  \\
  &
  \|\Delta q_i\|_2
  \leq
  \rho_i,
  \qquad i\in\{1,2\},
  \\
  &
  d_{\mathrm{obj/gnd}}(q_t+\Delta q)
  \geq
  -\epsilon,
  \\
  &
  d_{12}(q_t+\Delta q)+s
  \geq
  m.
\end{aligned}
\label{eq:rt-joint}
\end{equation}
Here, $E_U$ and $E_L$ preserve the upper interaction and lower support
structures, while $E_{\mathrm{hand}}$ and $E_{\mathrm{foot}}$ constrain active
hand contacts and foot support. Nominal and temporal terms preserve the GMR
prior and continuity. Joint limits, update bounds, and distance constraints maintain kinematic feasibility, reduce robot--object and robot--ground
penetration, suppress unintended foot motion, and preserve inter-robot
separation.

Cooperation is enforced through the upper and lower structures, active hand
anchors, and cross-robot constraint acting on the stacked dual-humanoid state
rather than independent motions. The
problem is solved frame by frame and warm-started from the previous frame.
The resulting trajectories serve as references for retargeting evaluation,
physics-based control, and real-world cooperative execution.

\begin{table*}[t]
\centering
\footnotesize
\setlength{\tabcolsep}{3.5pt}
\renewcommand{\arraystretch}{1.05}

\vspace{0.5em}
\caption{CoCarry-to-A3 retargeting results and component ablations.
Bold denotes the best reported value, with ties retained.}
\label{tab:cocarry-a3-results}

\begin{tabular}{lcccccc}
\hline

&
\multicolumn{1}{c}{\textbf{Interaction}} &
\multicolumn{2}{c}{\textbf{Clearance}} &
\multicolumn{2}{c}{\textbf{Support}} &
\multicolumn{1}{c}{\textbf{Output}}
\\[-0.2ex]

\cline{2-2}
\cline{3-4}
\cline{5-6}
\cline{7-7}

Method &
\shortstack{Contact\\preservation $\uparrow$} &
\shortstack{Penetration\\duration $\downarrow$} &
\shortstack{Penetration max.\\depth (m) $\downarrow$} &
\shortstack{Sliding\\duration $\downarrow$} &
\shortstack{Max. toe sliding\\velocity (m/s) $\downarrow$} &
\shortstack{Success\\rate (\%) $\uparrow$}
\\

\hline

GMR &
$0.7633$ &
$0.5840$ &
$0.1055$ &
\textbf{0.1458} &
$0.2978$ &
\textbf{100.00}
\\

OmniRetarget &
$0.9910$ &
$0.6640$ &
$0.0505$ &
$0.2047$ &
$0.1504$ &
$97.73$
\\

Ours w/o anchors &
$0.9994$ &
$0.6638$ &
$0.0501$ &
$0.2242$ &
\textbf{0.0603} &
\textbf{100.00}
\\

Ours w/o decomposition &
$0.9993$ &
$0.2026$ &
$0.0453$ &
$0.3600$ &
$0.0632$ &
$99.55$
\\

\hline

\textbf{Ours} &
\textbf{0.9995} &
\textbf{0.1925} &
\textbf{0.0451} &
$0.1913$ &
$0.0827$ &
\textbf{100.00}
\\

\hline
\end{tabular}

\end{table*}

\begin{figure*}[!t]
  \centering
  \includegraphics[width=0.96\textwidth]
  {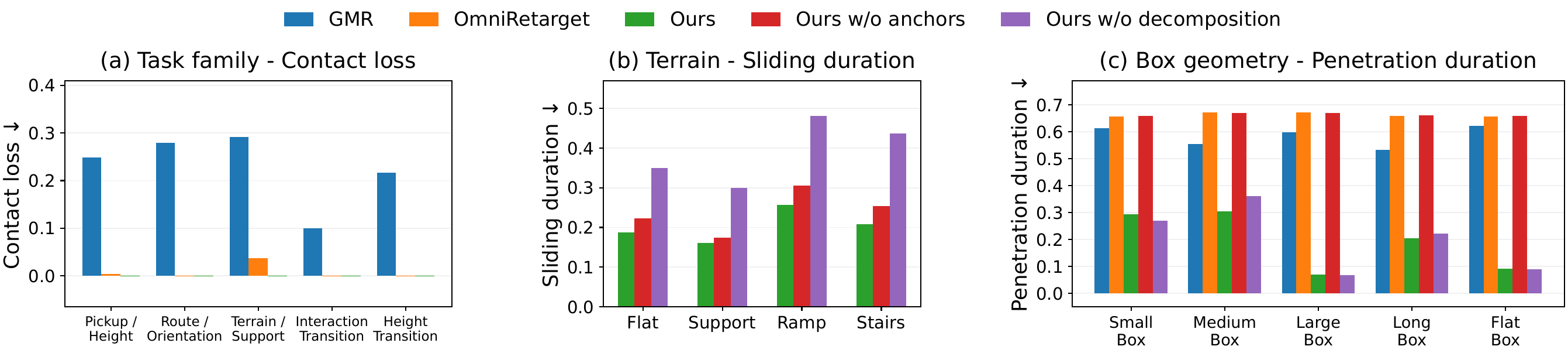}
  \caption{
  Condition-wise CoCarry-to-A3 results:
  (a) contact loss across task families,
  (b) sliding duration across terrain/support conditions, and
  (c) penetration duration across object geometries.
  Bars show means.
  Lower is better.
  }
  \label{fig:cocarry-conditionwise}
\end{figure*}

\section{Experiments}
\label{sec:experiments}

We evaluate CoCarry and the proposed Contact--Support-Structured Retargeting
at three levels: large-scale retargeting on CoCarry with AgiBot A3 humanoids,
external evaluation on CORE4D with Unitree G1 humanoids, and downstream
execution in simulation and on physical A3 humanoids. These experiments examine contact preservation, geometric validity, support
consistency, cross-dataset and cross-embodiment transfer, and downstream
usability.

\subsection{Experimental Setup and Metrics}
\label{sec:exp-setup}

The primary evaluation uses all 220 CoCarry sequences retargeted to two A3
humanoids at 50\,Hz. We compare GMR~\cite{gmr}, independently applied
OmniRetarget~\cite{omniretarget}, and our complete method. Two additional
variants remove the cooperation-aware hand anchors or replace the proposed
upper object-interaction and lower support structures with a unified
whole-body structure. All methods use identical source motions, object trajectories, scene assets,
and evaluation protocol.

We evaluate three aspects of the retargeted motion: hand--object interaction,
geometric clearance, and lower-body support. \emph{Contact preservation}
measures whether source hand--object contact remains after retargeting. A human hand is considered contact-relevant when it lies
within $0.10\,\mathrm{m}$ of the object surface while the object velocity
exceeds $0.03\,\mathrm{m/s}$; the corresponding robot hand must remain within
a signed surface distance of $0.02\,\mathrm{m}$. Because penetration can also
satisfy a proximity threshold, contact preservation is interpreted jointly
with the penetration metrics.

\emph{Penetration duration} measures the fraction of penetrating frames,
while \emph{penetration maximum depth} measures the severity of geometric
overlap. For lower-body support, \emph{sliding duration} measures the fraction
of support-like frames with unintended toe motion, and \emph{maximum
toe-sliding velocity} measures its severity. Human-toe horizontal velocity
below $0.01\,\mathrm{m/s}$ identifies support-like frames.

\emph{Retargeting success rate} is the percentage of sequences producing a
complete valid dual-humanoid trajectory.
Duration metrics are first computed per synchronized two-person sequence and
then averaged across sequences. All methods use the same thresholds and
aggregation protocol.

\subsection{CoCarry Retargeting Comparison}
\label{sec:exp-cocarry-results}

Table~\ref{tab:cocarry-a3-results} compares the proposed method with GMR and
OmniRetarget on the full CoCarry set. Our method achieves the highest contact
preservation ($0.9995$), the lowest penetration duration ($0.1925$), and the
lowest penetration maximum depth ($0.0451\,\mathrm{m}$), while retaining a
$100\%$ retargeting success rate.

Compared with GMR and OmniRetarget, penetration duration decreases from
$0.5840$ and $0.6640$ to $0.1925$, corresponding to relative reductions of
$67.0\%$ and $71.0\%$, respectively. GMR obtains a lower sliding duration,
but also exhibits substantially lower contact preservation, deeper
penetration, and a larger maximum toe-sliding velocity. Thus, low foot motion
alone does not imply a geometrically valid cooperative reference when shared
grasp or object clearance is violated.

Figure~\ref{fig:cocarry-conditionwise} further examines whether these gains
persist across different conditions. Across the five task families, Ours
maintains contact loss below $0.001$, whereas GMR ranges from $0.1002$ to
$0.2911$ and OmniRetarget reaches $0.0366$ for terrain and support motions.
Across flat ground, support surfaces, ramps, and stairs, Ours obtains the
lowest sliding duration among the three cooperative variants, while the
unified-mesh ablation is consistently the least stable. Across all five
object geometries, Ours remains below both generic baselines in penetration
duration, whereas removing the anchors returns penetration to
$0.6580$--$0.6706$, close to OmniRetarget. These trends show that the improvements are not confined to a particular
task, terrain, or object geometry.

\subsection{Component Analysis}
\label{sec:exp-ablation}

The two ablations in Table~\ref{tab:cocarry-a3-results} isolate the roles of
the cooperation-aware hand anchors and the contact--support structured
representation.

Removing the hand anchors leaves the threshold-based contact score nearly
unchanged ($0.9994$ vs.\ $0.9995$), but increases penetration duration from
$0.1925$ to $0.6638$. This highlights a limitation of proximity-based contact:
a hand may remain close to the object while occupying an incompatible surface
region or orientation. The explicit anchors provide additional geometric
guidance toward compatible four-hand grasp configurations.

Replacing the upper object-interaction and lower support structures with a
unified whole-body representation increases sliding duration from $0.1913$
to $0.3600$. The proposed decomposition therefore reduces sliding duration by
$46.9\%$ while retaining comparable contact and penetration quality.

The complete method produces valid trajectories for all 220 CoCarry
sequences. GMR also reaches $100\%$ output success, while OmniRetarget and
the unified whole-body variant achieve $97.73\%$ and $99.55\%$,
respectively.

\begin{figure*}[!t]
  \centering
  \includegraphics[width=0.96\textwidth]
  {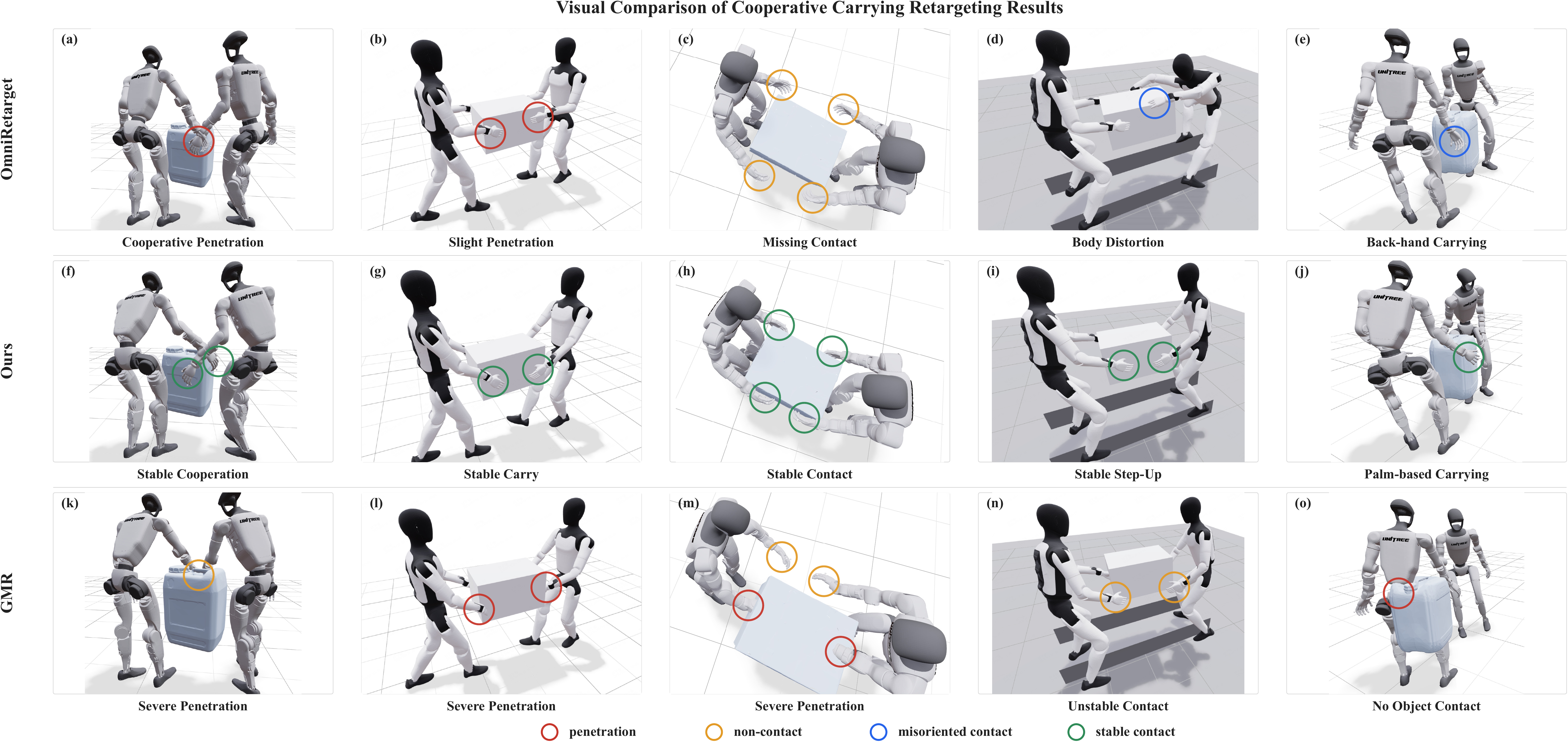}
  \caption{
  Visual comparison across CoCarry-to-A3 and CORE4D-to-G1 settings.
  Rows correspond to OmniRetarget, Ours, and GMR, while columns show matched
  examples illustrating penetration, missing contact, body distortion, and
  incorrect hand orientation. Colored markers indicate penetration (red),
  missing contact (orange), misoriented contact (blue), and correctly
  preserved contact (green).
  }
  \label{fig:qualitative-comparison}
\end{figure*}

\subsection{Cross-Dataset and Cross-Embodiment Evaluation}
\label{sec:exp-core4d}

We further evaluate the method on CORE4D~\cite{core4d} using two Unitree G1
humanoids. To separate trajectory quality from method-specific output
failures, we construct a matched shared-success subset containing 200
sequences for which GMR, OmniRetarget, and our method all produce valid
outputs. The subset covers multiple manipulated-object categories, including
bucket, box, desk, chair, and stick.

CORE4D human motions, object trajectories, and scene assets are converted to
the same retargeting representation as CoCarry. The robot model,
semantic kinematic correspondence, interaction geometry, and optimization
configuration are adapted to the Unitree G1 embodiment.

\begin{table}[t]
\centering
\scriptsize
\setlength{\tabcolsep}{1.5pt}
\renewcommand{\arraystretch}{0.92}
\caption{CORE4D-TO-G1 RETARGETING RESULTS.}
\label{tab:core4d-g1-results}
\begin{tabular}{lccccc}
\hline
Method &
\shortstack{Contact\\pres. $\uparrow$} &
\shortstack{Pen.\\duration $\downarrow$} &
\shortstack{Max.\\depth (m) $\downarrow$} &
\shortstack{Sliding\\duration $\downarrow$} &
\shortstack{Max. toe\\speed (m/s) $\downarrow$} \\
\hline
GMR & $0.6908$ & $0.8386$ & $0.0802$ & $0.1519$ & $0.0115$ \\
OmniRetarget & $0.7337$ & {\boldmath$0.0008$} & {\boldmath$0.0004$} & $0.0242$ & $0.0019$ \\
Ours & {\boldmath$0.9906$} & $0.0167$ & $0.0039$ & {\boldmath$0.0000$} & {\boldmath$0.0000$} \\
\hline
\end{tabular}
\end{table}

As shown in Table~\ref{tab:core4d-g1-results}, our method achieves
$0.9906$ contact preservation, compared with $0.6908$ for GMR and
$0.7337$ for OmniRetarget. Sliding duration and maximum toe-sliding velocity
are both reduced to zero on the matched subset, indicating strong support
stability after changing both the source dataset and robot embodiment.

OmniRetarget obtains lower penetration duration and maximum depth than our
method, revealing a contact--clearance trade-off rather than uniform
improvement across every metric. Our formulation more strongly preserves
cooperative grasp while introducing a small amount of shallow penetration;
both penetration measures nevertheless remain substantially below those of
GMR.

Figure~\ref{fig:qualitative-comparison} provides qualitative examples from
both CoCarry-to-A3 and CORE4D-to-G1 settings. Across both embodiments,
independent retargeting can produce penetration, missing contact, body
distortion, or incorrect hand orientation, whereas our formulation more
consistently preserves palm-facing contact and compatible cooperative grasp
geometry.

\subsection{Simulation and Real-World Execution}
\label{sec:exp-execution}

\begin{figure}[!t]
  \centering
  \setlength{\tabcolsep}{1pt}
  \renewcommand{\arraystretch}{1.0}
  \begin{tabular}{@{}cccc@{}}

    \multicolumn{4}{c}{
      \footnotesize\textbf{(a)} Physics simulation
    }\\[0.2em]

    \includegraphics[width=0.24\columnwidth]{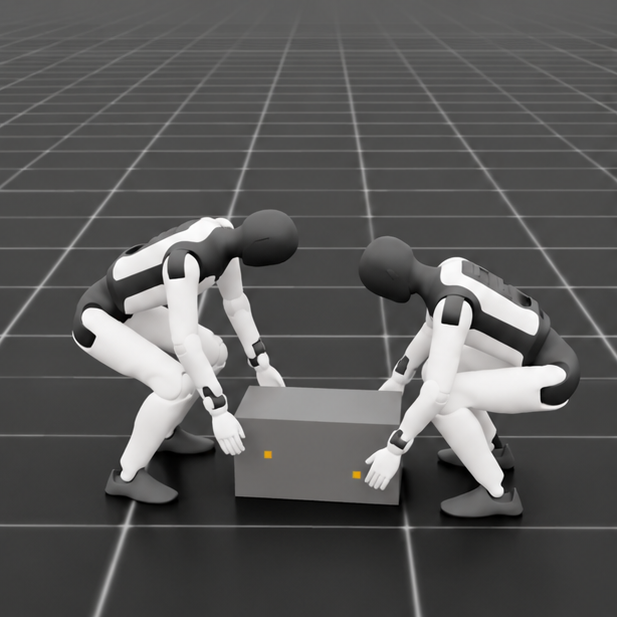} &
    \includegraphics[width=0.24\columnwidth]{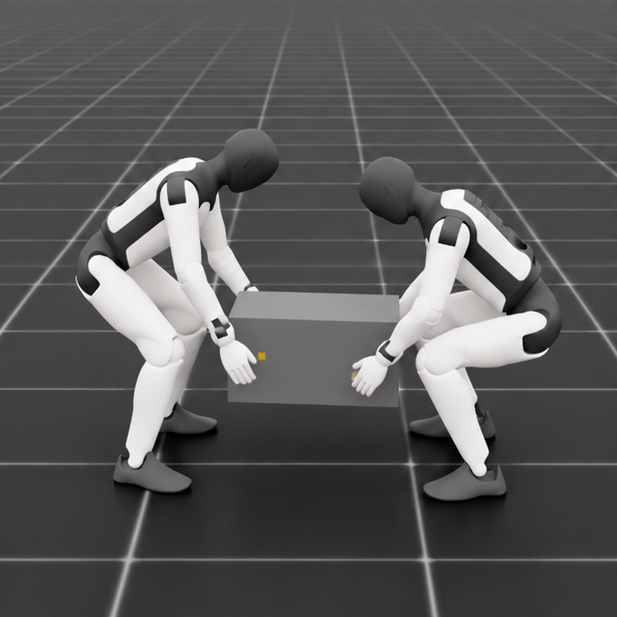} &
    \includegraphics[width=0.24\columnwidth]{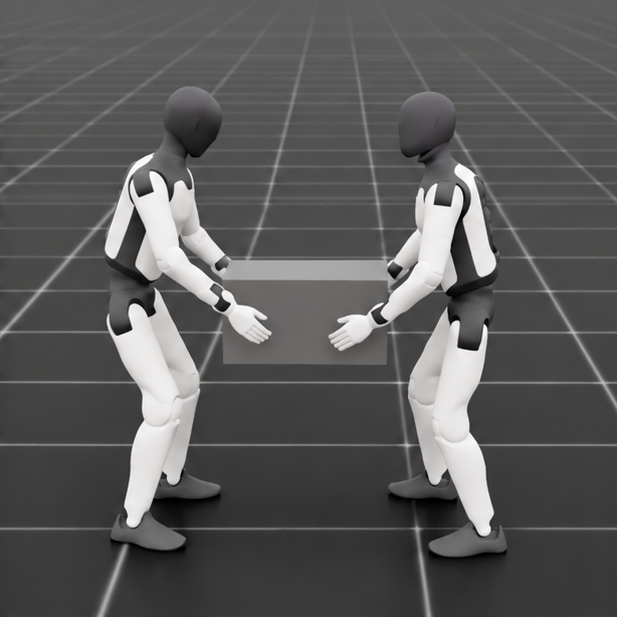} &
    \includegraphics[width=0.24\columnwidth]{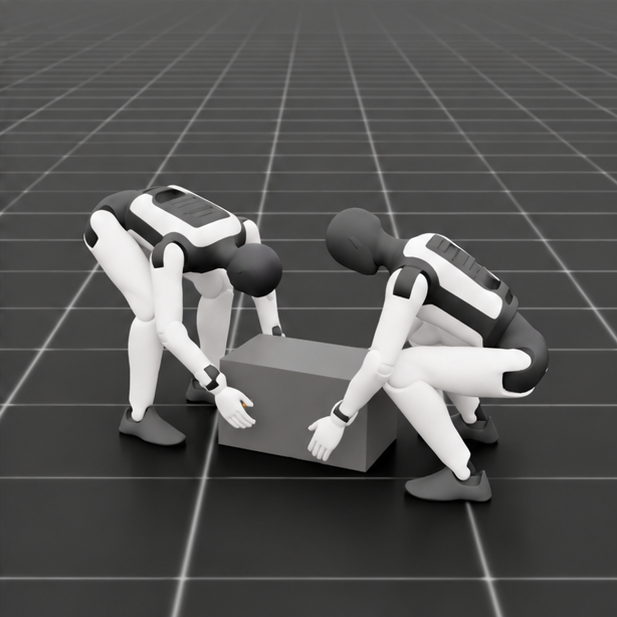} \\

    \scriptsize Approach &
    \scriptsize Lift &
    \scriptsize Carry &
    \scriptsize Place \\[0.3em]

    \multicolumn{4}{c}{
      \footnotesize\textbf{(b)} Real-world execution
    }\\[0.2em]

    \includegraphics[width=0.24\columnwidth]{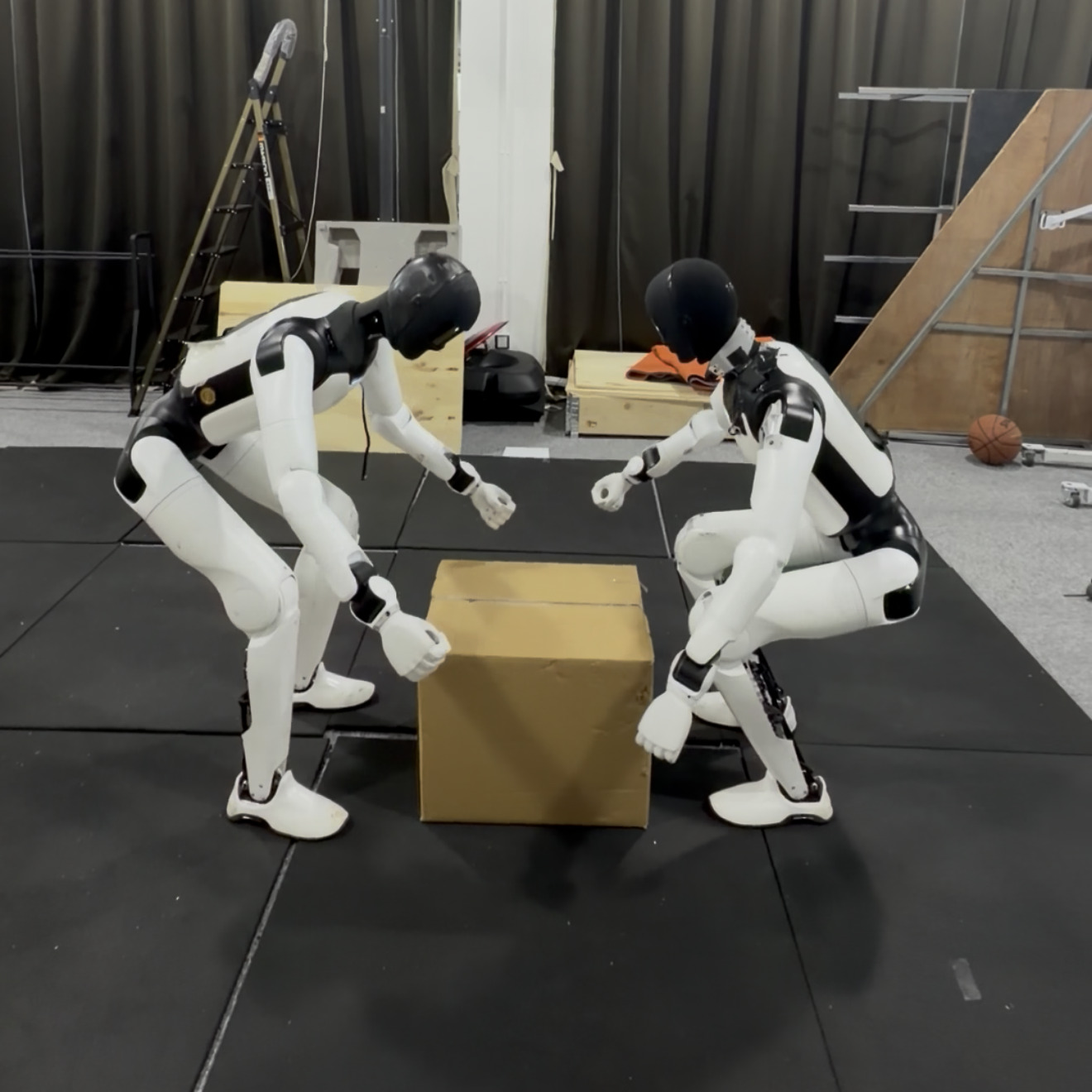} &
    \includegraphics[width=0.24\columnwidth]{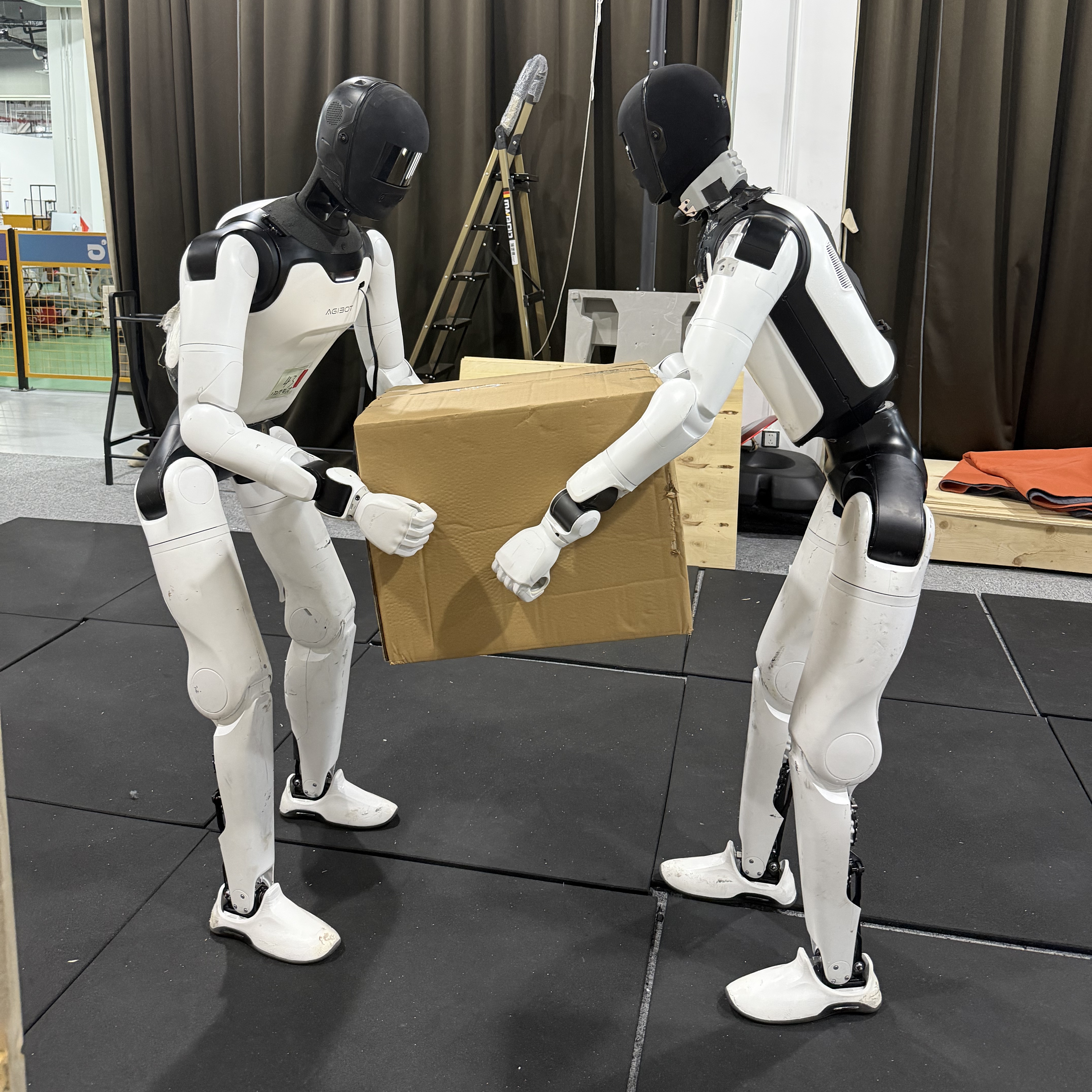} &
    \includegraphics[width=0.24\columnwidth]{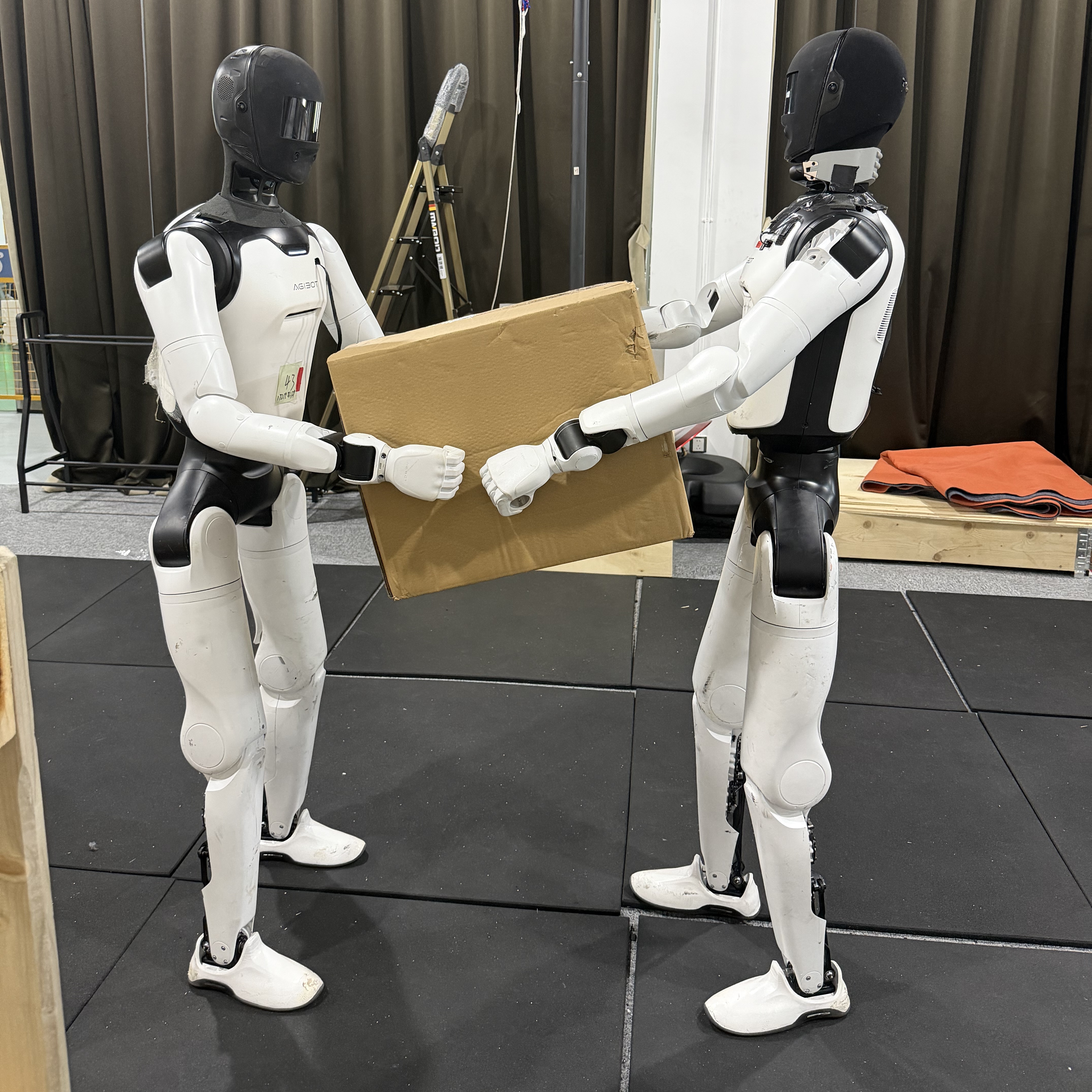} &
    \includegraphics[width=0.24\columnwidth]{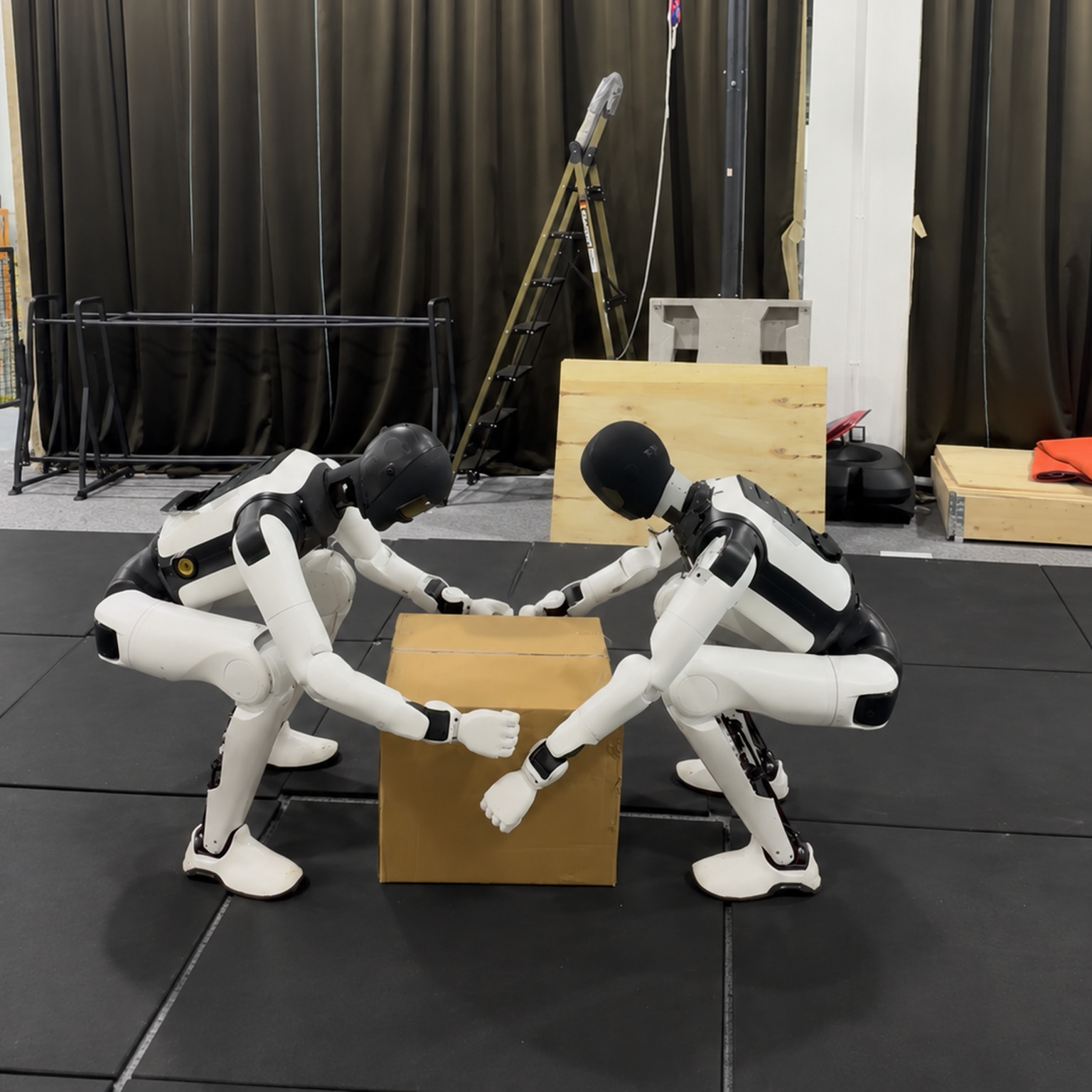} \\

    \scriptsize Approach &
    \scriptsize Lift &
    \scriptsize Carry &
    \scriptsize Place

  \end{tabular}

  \caption{
  Representative downstream execution using two A3 humanoids.
  (a) Approach, lift, carry, and place keyframes from physics simulation with
  a free shared object.
  (b) Representative keyframes from real-world cooperative carrying
  experiments on two physical A3 humanoids.
  }
  \label{fig:a3-execution}
  \vspace{-1ex}
\end{figure}

Finally, we evaluate whether the generated references are usable by
downstream humanoid controllers. Retargeted CoCarry motions are
used for motion-imitation training with two A3 humanoids. Unlike the
retargeting stage, where the object trajectory is prescribed, the simulation
uses a free shared object whose motion emerges from coordinated robot
interaction.

In physics simulation, the learned policies reproduce cooperative behaviors
including approach, lifting, transport, placement, and release. We further
deploy the resulting controllers on two physical A3 humanoids and validate the
pipeline through multiple real-world cooperative carrying motions.

Figure~\ref{fig:a3-execution} shows representative keyframes from simulation
and real-world execution. These experiments provide downstream physical validation rather than a
dataset-wide control benchmark, showing that the references are usable for
simulated and real-world dual-humanoid control.

\section{Conclusion}
\label{sec:conclusion}

We presented \textbf{CoCarry}, a cooperative-carrying dataset that preserves
synchronized two-person motion, shared-object trajectories, and time-varying
interaction priors, together with \textbf{Contact--Support-Structured
Retargeting} for converting human cooperative motion into dual-humanoid
references. The proposed method combines cooperation-aware hand anchors with
complementary upper object-interaction and lower support structures, jointly
refining both humanoids around the same prescribed object trajectory.

Experiments on all 220 CoCarry sequences demonstrate strong cooperative
contact preservation and reduced penetration on two A3 humanoids, while
external evaluation on CORE4D with Unitree G1 humanoids provides evidence of
cross-dataset and cross-embodiment transfer. The generated references are
further validated through physics simulation and real-world cooperative
carrying on two physical A3 humanoids.

Future work will extend the framework to broader cooperative manipulation
tasks, additional humanoid platforms, and larger-scale real-world evaluation.

\balance

\bibliographystyle{IEEEtran}
\bibliography{references}

\end{document}